\pdfoutput=1

\documentclass[11pt]{article}

\usepackage{acl}

\usepackage{booktabs}
\usepackage{times}
\usepackage{latexsym}
\usepackage{xspace}
\usepackage{graphicx}
\usepackage{subcaption}
\usepackage{mathtools}
\usepackage[T1]{fontenc}

\usepackage[utf8]{inputenc}

\usepackage{microtype}
\newcommand{\dataset}{\textsc{SummScreen}\xspace}
\newcommand{\tvmegasite}{TVMegaSite\xspace}
\newcommand{\foreverdream}{ForeverDreaming\xspace}

\newcommand{\tvmegasiteshort}{TMS\xspace}
\newcommand{\foreverdreamshort}{FD\xspace}

\newenvironment{enumeratesquish}{\begin{list}{\addtocounter{enumi}{1}\labelenumi}{\setlength{\itemsep}{0em}\setlength{\labelwidth}{0.5em}\setlength{\leftmargin}{\labelwidth}\addtolength{\leftmargin}{\labelsep}}}{\end{list}\setcounter{enumi}{0}}

\title{screenplay summarization}
\title{SummScreen: A Dataset for Screenplay Summarization} 
\title{Challenges in Summarizing Screenplays}
\title{To Be Continued: Abstractive Screenplay Summarization}
\title{Must Read TV: Abstractive Screenplay Summarization}
\title{SummScreen: A Challenge Dataset for Abstractive Screenplay Summarization}
\title{SummScreen: A Dataset for Abstractive Screenplay Summarization} 

\author{Mingda Chen$^{1}$ \quad Zewei Chu\thanks{~~Work done while the author was at the University of Chicago.} \quad Sam Wiseman$^{2}$\thanks{~~Work done while the author was at Toyota Technological Institute at Chicago.} \quad Kevin Gimpel$^{1}$ \\
$^{1}$Toyota Technological Institute at Chicago, IL, USA\\
$^{2}$Duke University, NC, USA \\
  \texttt{\{mchen,kgimpel\}@ttic.edu},~\texttt{swiseman@cs.duke.edu},~\texttt{zeweichu@gmail.com}
}

\begin{document}
\maketitle
\begin{abstract}
We introduce \dataset, a summarization dataset comprised of pairs of TV series transcripts and human written recaps. 
The dataset provides a challenging testbed for abstractive summarization for several reasons. 
Plot details are often expressed indirectly in character dialogues and may be scattered across the entirety of the transcript. These details must be found and integrated to form the succinct plot descriptions in the recaps. 
Also, TV scripts contain content that does not directly pertain to the central plot but rather serves to develop characters or provide comic relief. This information is rarely contained in recaps. 
Since characters are fundamental to TV series, we also propose two entity-centric evaluation metrics. 
Empirically, we characterize the dataset by evaluating several methods, including neural models and those based on nearest neighbors. 
An oracle extractive approach outperforms all benchmarked models according to automatic metrics, showing that the neural models are unable to fully exploit the input transcripts.
Human evaluation and qualitative analysis reveal that our non-oracle models are competitive with their oracle counterparts in terms of generating faithful plot events 
and can benefit from better content selectors. Both oracle and non-oracle models  generate unfaithful facts, suggesting future research directions.\footnote{\dataset is available at \url{https://github.com/mingdachen/SummScreen}}

\end{abstract}

\section{Introduction}

\begin{figure}[t]
    \centering
    \includegraphics[scale=0.5]{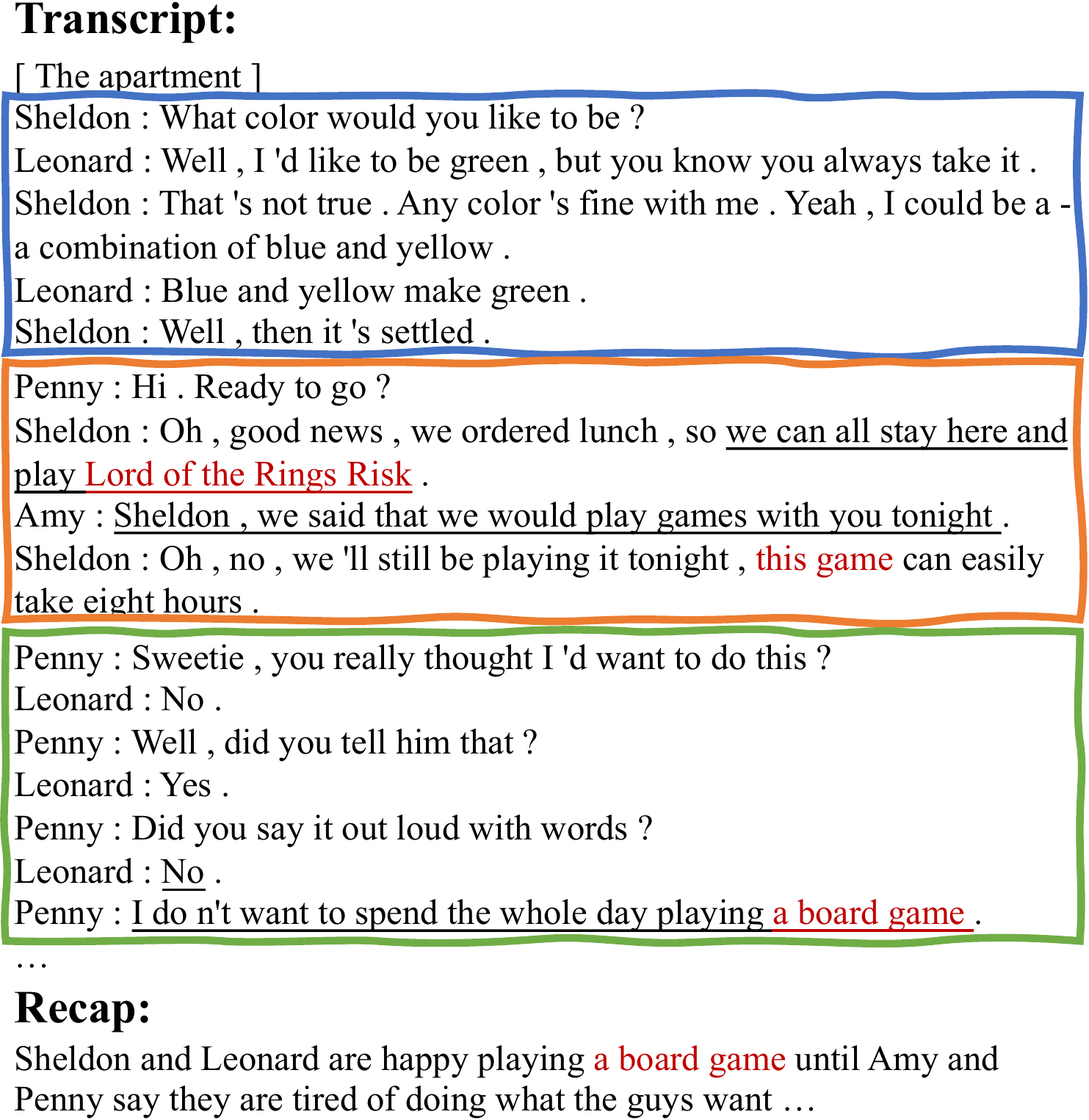}
    \caption{Excerpts from an example from \dataset. The transcript and recap are from the TV show ``The Big Bang Theory''. 
    Generating this sentence in the recap requires discerning the characters' feelings (clues in the transcript are underlined) about playing the board game (references are shown in red).  %
    Colored boxes indicate utterances belonging to the same conversations. }
\vspace{-1.3em}
    \label{fig:intro_example}
\end{figure}

Abstractive summarization aims to produce a summary that concisely expresses key points of the input document 
rather than merely extracting pieces of it. %
Existing datasets are constructed from various domains, such as news \cite{linguistic2008evan,teaching2015hermann,rush-etal-2015-neural,narayan-etal-2018-dont,grusky-etal-2018-newsroom}, online forums \cite{volske-etal-2017-tl}, meeting dialogues \cite{icsi2003,carletta2005ami}, and webpages \cite{10.1145/3366423.3380206}. However, few datasets exist for abstractive summarization of narrative text, which focuses on entities and dialogue among entities, with plot details often communicated indirectly via dialogue. In this work, we build \dataset, an abstractive summarization dataset combining TV series transcripts %
and episode recaps. Figure \ref{fig:intro_example} shows an example from \dataset.

Several aspects of \dataset make it a challenging testbed for abstractive summarization. First, the relationship between character dialogue and plot details is not straightforward. Plot events are often expressed indirectly in dialogue, and dialogue contains other information that is not directly relevant to the plot, such as character development and humor. Also, a typical episode has multiple subplots that proceed in parallel, with consecutive scenes often describing different subplots.  
Solving \dataset requires drawing information from utterances across a wide range of the input and integrating the information to form concise plot descriptions.
Moreover, since actual TV episodes ground their scripts with audio-visual accompaniment, many details may be omitted from the transcript itself. This omission of details and the other challenging aspects mentioned above have inspired research into other NLP tasks on TV show transcripts, such as entity tracking \cite{chen-choi-2016-character,choi-chen-2018-semeval} and coreference resolution \cite{chen-etal-2017-robust,zhou-choi-2018-exist}.

Another prominent characteristic of TV series transcripts is their focus on characters. To reflect this aspect, we propose two entity-centric metrics to evaluate the quality of generated plot summaries. One is based on bags of characters, which measures the overlap of the characters that appear in both the generated and reference recaps. The other metric measures character relations: the overlap of cooccurrences of character pairs in generations and recaps.

We empirically evaluate several types of methods on \dataset.  We consider nearest neighbor models, which look up similar transcripts or recaps, neural abstractive summarization models, and hybrid models, which use the nearest neighbor models as content selectors followed by abstractive summarization. Oracle extractive approaches outperform all models on all the automatic metrics.
These results suggest that the benchmarked methods are unable to fully exploit the input transcripts and that improving content selection may be a promising research direction.

Human evaluations show that our non-oracle hybrid models are competitive with their oracle counterparts in terms of generating faithful plot events. Hybrid models may be promising approaches for future research. Qualitative analysis shows that neural models tend to generate generic summaries, hybrid models can benefit from better content selection, and hybrid models sometimes generate unfaithful details. 

\section{Related Work}

There has been prior work on \emph{extractive} screenplay summarization \cite{gorinski-lapata-2015-movie,papalampidi-etal-2020-screenplay}, and analyzing crime drama \cite{frermann-etal-2018-whodunnit}. The majority of TV show transcripts are dialogues, relating our work to prior work on dialogue and meeting summarization. 
Relevant datasets have been studied for medical dialogues \cite{joshi-etal-2020-dr,krishna2020generating}, chitchat (SAMSum; \citealp{gliwa-etal-2019-samsum}), podcasts \cite{clifton-etal-2020-100000}, meetings (AMI; \citealp{carletta2005ami}; ICSI; \citealp{icsi2003}; QMSum; \citealp{zhong2021qmsum}), livestreams (StreamHover; \citealp{cho-etal-2021-streamhover}), online forums (ForumSum; \citealp{khalman-etal-2021-forumsum-multi}) and news interviews (MediaSum; \citealp{zhu2021mediasum}).

There have been attempts in summarizing long-form text (other than screenplays), such as books \cite{mihalcea-ceylan-2007-explorations}, scientific articles (PubMed and arXiv; \citealp{cohan-etal-2018-discourse}), multiple news articles (Multi-News; \citealp{fabbri-etal-2019-multi}),  opinionated text (RottenTomatoes; \citealp{wang-ling-2016-neural}), patents \cite{sharma-etal-2019-bigpatent}, TV show stories (TVRecap; \citealp{chen2021tvrecap}) and (extractive summarization of) chapters of novels \cite{ladhak-etal-2020-exploring}. More detailed discussion on the differences between these datasets and \dataset is in the next section.

Recently there have been efforts on adapting resources for TV shows for different tasks, including question answering \cite{ma-etal-2018-challenging,yang-choi-2019-friendsqa}, speaker identification \citep{ma-etal-2017-text}, sarcasm detection \citep{joshi-etal-2016-harnessing}, emotion detection \citep{zahiri2017emotion,hsu-ku-2018-socialnlp}, character relation extraction \cite{yu-etal-2020-dialogue}, and story generation \cite{chen2021tvrecap}.

\section{\dataset}

An instance in \dataset contains a transcript from TV series and its corresponding recap. The transcripts consist of dialogue utterances with speaker names, and descriptions of scenes or character actions. The recaps are human-written summaries of the corresponding transcripts. Figure \ref{fig:intro_example} shows an example in \dataset from the TV show ``The Big Bang Theory''. The transcript documents a dialogue involving four characters (Sheldon, Leonard, Penny, and Amy) about playing a board game, and the recap summarizes the dialogue into sentences. 

\subsection{Dataset Construction}\label{sec:dataset_construct}

\begin{table}[t]
    \centering\small
    \setlength{\tabcolsep}{5pt}
    \begin{tabular}{|l|r|r|r|r|r|r|}\hline
         & uni. & bi. & tri. & four. & src. & tgt. \\\hline
        \multicolumn{7}{|c|}{\dataset}\\\hline
        \foreverdreamshort & 81.6 & 29.9 & 5.6 & 1.3 & 7.6k & 113.7 \\
        \tvmegasiteshort & 86.5  & 34.1 & 6.9 & 2.1 & 6.4k & 380.6 \\\hline
        \multicolumn{7}{|c|}{Other summarization datasets}\\\hline
        XSum\textsuperscript{\dag} & 64.2 & 16.6 & 4.5 & 1.5 & 431.1 & 23.3 \\
        CNNDM\textsuperscript{\S} & 80.5 & 43.1 & 25.6 & 17.2 & 810.6 & 56.2\\
        MNews\textsuperscript{\S} & 82.2 & 42.9 & 24.3 & 17.7 & 2.1k & 264.7 \\\hline
    \end{tabular}
    \caption{Fraction (\%) of n-grams in the \textit{output summaries} that also appear in the inputs, and the average numbers of tokens for the inputs and outputs. Datasets with smaller fractions of overlapping n-grams tend to favor abstractive summarization approaches. Results marked by \dag\xspace and \S\xspace are from \citet{narayan-etal-2018-dont} and \citet{fabbri-etal-2019-multi} respectively.}
    \label{tab:overlap_ratio_summarization_dataset_compare}
\end{table}

We use two sources to construct \dataset: 
The TV MegaSite, Inc. %
(\tvmegasiteshort)\footnote{\url{http://tvmegasite.net/}} 
and \foreverdream (\foreverdreamshort),\footnote{\url{transcripts.foreverdreaming.org}} both 
of which provide
community-contributed transcripts. As \foreverdreamshort does not provide recaps, we obtain recaps 
of \foreverdreamshort shows
from Wikipedia and TVMaze.\footnote{\url{www.tvmaze.com}, an online TV database curated by TV fans. } 
To ensure dataset quality of \dataset, we filter out instances based on two criteria. First, the overlap ratio of TV show characters appearing in the recap and its transcript should be higher than 85\%. We use this criterion to ensure that the alignments between recaps and transcripts are correct. 
Second,
the number of transcript lines that have speaker information (``character utterances'') should be larger than 100. We use this criterion to eliminate transcripts that are essentially subtitles, i.e., utterances without speaker information. In practice, for each transcript line, if a colon symbol appears in the first 8 tokens and there exists at least one character name in front of the colon symbol, we will count it as a character utterance. We note that \foreverdreamshort and \tvmegasiteshort do not have overlapping TV series.

In Table~\ref{tab:overlap_ratio_summarization_dataset_compare}, we compute n-gram overlap ratios between recaps and transcripts for measuring the abstractiveness of \dataset. From the results, We find that despite \dataset has longer summaries, its fraction of overlapping four-gram is comparable to XSum \cite{narayan-etal-2018-dont} which is known for abstractiveness, suggesting that \dataset favors abstractive approaches.

\begin{table}[t]
    \centering\small
\begin{tabular}{|l|r|r|}\hline
& \foreverdreamshort & \tvmegasiteshort \\\hline
number of shows & 88 & 10 \\
number of episodes & 4348 & 22503 \\
min. \# episodes per show & 1 & 168 \\
max. \# episodes per show & 568 & 3784 \\
median \# episodes per show & 9.0 & 1973.5 \\
avg. \# episodes per show & 49.4 & 2250.0 \\
\hline
avg. \# tokens in recaps & 113.7 & 380.6 \\
avg. \# tokens in transcripts & 7605.4 & 6420.7 \\
avg. \# lines in transcripts & 447.6 & 360.8 \\
avg. \# char. utterances in transcripts & 330.7 & 327.0 \\
avg. \# uniq. char. in recaps & 5.8 & 14.3 \\
avg. \# uniq. char. in transcripts & 20.6 & 29.8 \\\hline
\end{tabular}
    \caption{Detailed dataset statistics for \dataset.}
    \label{tab:dataset_stats}
\end{table}

\begin{figure}[t]
    \centering
    \begin{subfigure}[t]{0.25\textwidth}
    \centering
    \small
    \begin{tabular}{|l|r|}%
    \hline
      Genre   & Count \\\hline
    Drama & 65\\
    Romance & 24\\
      Comedy & 23\\
     Crime & 18\\
      Action & 15 \\
    Science-Fiction & 12\\
      Adventure & 9 \\
    Supernatural & 9\\
    Mystery & 8\\
    Thriller & 5\\
    Family & 5\\
    Medical & 5\\
    Fantasy & 4\\
    Horror & 4\\
    History & 3\\
    Sports & 3\\
    Western & 3\\
      Children & 2\\
    Legal & 2\\
    Espionage & 1\\
    Music & 1\\%\hline
    \hline
    \end{tabular}
    \end{subfigure}%
    \begin{subfigure}[t]{0.25\textwidth}
    \centering
    \small
    \begin{tabular}{|l|r|}\hline
       Genre  & Count \\\hline
        Drama & 10 \\
       Romance & 6 \\
       Family & 4 \\
       Medical & 1 \\
        \hline
    \end{tabular}
    \end{subfigure}
    \caption{Left: TV show genres from \foreverdream. Right: TV show genres from \tvmegasite. 
    }
    \label{fig:dataset_genres}
\end{figure}

Table \ref{tab:dataset_stats} shows data statistics and Figure \ref{fig:dataset_genres} shows the genres of the TV shows from the two sources.\footnote{The genre information is from TVMaze where a TV show may correspond to multiple genres.}
When computing the number of unique characters in TV shows,
we first collect the character names from TVMaze and the named entities\footnote{We use the named entity recognizer from spaCy \cite{spacy2}.} preceding the colon symbols in transcripts. We then 
perform string matching to obtain numbers of TV show characters in recaps and transcripts. From these two tables, we observe
that \foreverdreamshort and \tvmegasiteshort are different in many aspects. First, \foreverdreamshort covers more diverse genres than \tvmegasiteshort. This is partly due to the fact that TV shows 
from \tvmegasiteshort are soap operas. 
Second, transcripts from \foreverdreamshort are longer, which is caused by the fact that the transcripts from \foreverdreamshort tend to have more descriptions about environments or character actions, whereas the ones from \tvmegasiteshort are mostly made up of dialogue (see Table~\ref{tab:dataset_stats}). Third, recaps from \foreverdreamshort are shorter whereas recaps from \tvmegasiteshort seek to cover more details. Fourth, writing styles are more diverse in \foreverdreamshort than those in \tvmegasiteshort. In light of these differences, we treat \foreverdreamshort and \tvmegasiteshort as different datasets in the following experiments.

\begin{table}[t]
    \centering\small
\begin{tabular}{|l|ccc|} %
\hline
\foreverdream & Train & Dev & Test \\ \hline
\# shows & 66 & 78 & 81 \\ %
\# episodes & 3673 & 338 & 337 \\%\hline
\hline
\tvmegasite & Train & Dev & Test \\ \hline
\# shows & 10 & 10 & 10 \\ %
\# episodes & 18915 & 1795 & 1793 \\ %
\hline
\end{tabular}
    \caption{Statistics of train/dev/test splits for \foreverdream and \tvmegasite. }
    \label{tab:split_stat}
\end{table}

We create train/dev/test splits for \foreverdreamshort and \tvmegasiteshort by ensuring the ratio to be roughly 10:1:1, and filter out instances in the dev/test splits if the reference texts are shorter than 30 word tokens. The statistics of the splits are shown in Table \ref{tab:split_stat}.

\begin{table*}[t]
    \centering\small
    \begin{tabular}{|l|r|r|r|r|c|}\hline
         & \# instances & \# tokens (input) & \# tokens (summary) & \# speakers & Domain \\\hline
        Multi-News & 56.2k & 2103.5 & 264.7 & - & News \\
        RottenTomatoes & 3.7k & 2124.7 & 22.2 & - & Reviews \\
        arXiv  & 215k & 4938.0 & 220.0 & - & Science \\
        PubMed & 113k & 3016.0 & 203.0 & - & Science \\
        GovReport & 19.5k & 9409.4 & 553.4 & - & Government Reports \\
        TVRecap & 29.0k & 1868.7 & 221.6 & - & Television Series \\
        \hline
        SAMSum & 16.4k & 83.9 & 20.3 & 2.2 & Chitchat \\
        ForumSum & 4.1k & 303.5 & 36.0 & 6.7 & Forum Messages \\
        MediaSum & 463.6k & 1553.7 & 14.4 & 6.5 & News Interviews \\
        AMI  & 137 & 4757.0 & 322.0 & 4.0 &  Meetings \\
        ICSI  & 59  & 10189.0 & 534.0 & 6.2 &  Meetings \\
        QMSum & 1.8k & 9069.8 & 69.6 & 9.2 & Meetings \\
        \dataset & 26.9k & 6612.5 & 337.4 & 28.3 & Television Series         \\\hline
    \end{tabular}
    \caption{Statistics for datasets focusing on abstractive summarization for long-form text or dialogue. The numbers are averaged over instances. We omit number of speakers for datasets that do not contain dialogue. \dataset combines long source inputs, large numbers of speakers, and a moderate number of instances.
    }
    \label{tab:compare_dataset}
\end{table*}

\subsection{Dataset Comparison}

We compare \dataset to other abstractive dialogue summarization datasets in Table~\ref{tab:compare_dataset}. \dataset differs from other datasets in several ways:
\begin{enumeratesquish} %
\item %
Compared to recently proposed large-scale dialogue summarization datasets (i.e., SAMsum and MediaSUM), \dataset has longer source inputs. 
\item %
Compared to other dialogue summarization datasets, \dataset has larger numbers of speakers per instance. The TV series genre focuses on narrative, which is typically entity-centric and can include multiple parallel subplots in a single episode. 
\item %
Compared to AMI, ICSI and QMSum, which are long-input meeting summarization datasets, \dataset has far more instances. 
\item
Unlike most of the other datasets, \dataset contains many episodes of a single show (e.g., more than 3k episodes for \tvmegasiteshort). This episodic structure could be used to model character arcs, the evolution of character personality traits and character relationships over episodes, among others.
\end{enumeratesquish}
Properties (1) and (2) above make extracting information from transcripts more challenging than other datasets. The third property means that \dataset is large enough to train and evaluate neural methods. 

The Spotify Podcast Dataset \cite{clifton-etal-2020-100000} and StreamHover  \cite{cho-etal-2021-streamhover} are similar to \dataset in that they contain transcribed speech and summaries. However, the transcriptions are obtained automatically and therefore contain errors.\footnote{For this reason, we do not include their statistics in Table~\ref{tab:compare_dataset}.} The datasets therefore involve speech processing (or at least handling speech recognition errors) compared to \dataset, which has human-written transcripts. 

Since MediaSum is constructed from news transcripts, it is the most similar dataset in Table~\ref{tab:compare_dataset} to \dataset. However, the summaries in MediaSum are twenty times shorter than those in \dataset, and the average number of speakers per instance is only a quarter of that in \dataset. Furthermore, our results in Sec.~\ref{sec:main_results} indicate that our dataset is much harder than MediaSum as the pretrained models perform worse on our dataset than on MediaSum according to automatic metrics. More detailed analysis is in the next subsection.

\subsection{Dataset Challenges}

In this subsection, we qualitatively analyze the challenging aspects of \dataset.
Since the transcripts focus on dialogue among characters, along with limited descriptions of scenes and actions, it leads to the challenge that plot information is not stated explicitly but rather only implied in the dialogue. For example, the transcript in Figure~\ref{fig:intro_example} does not explicitly describe what Sheldon and Leonard are playing. However, it is implied by Sheldon when he mentions playing ``Lord of the Rings Risk,'' and later by Penny when she says that she does not ``want to spend the whole day playing a board game.''

\begin{figure*}[t]
    \centering
    \begin{subfigure}{0.5\textwidth}
    \centering
    \includegraphics[scale=0.55]{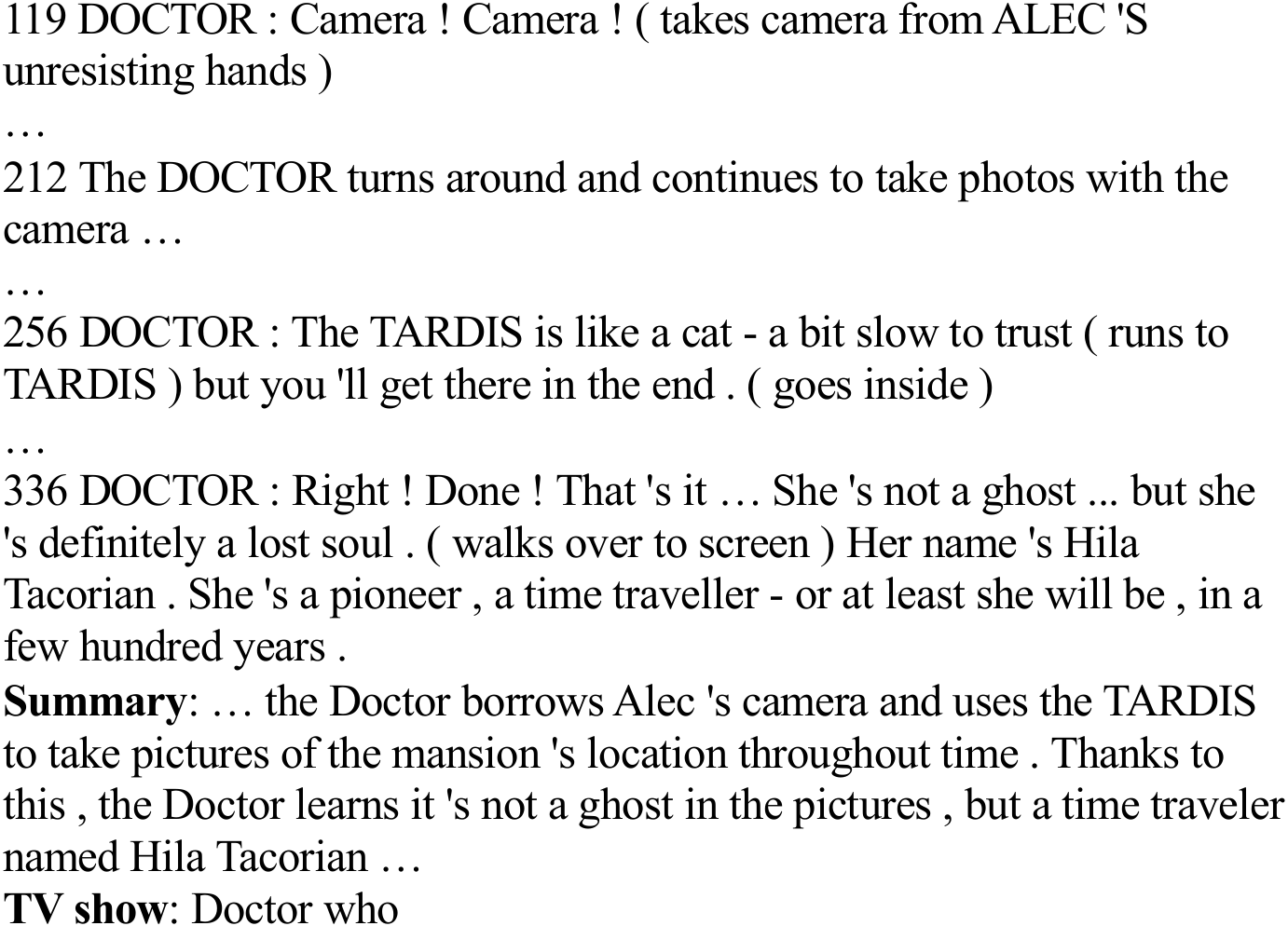}
    \end{subfigure}%
    \begin{subfigure}{0.5\textwidth}
    \centering
    \includegraphics[scale=0.55]{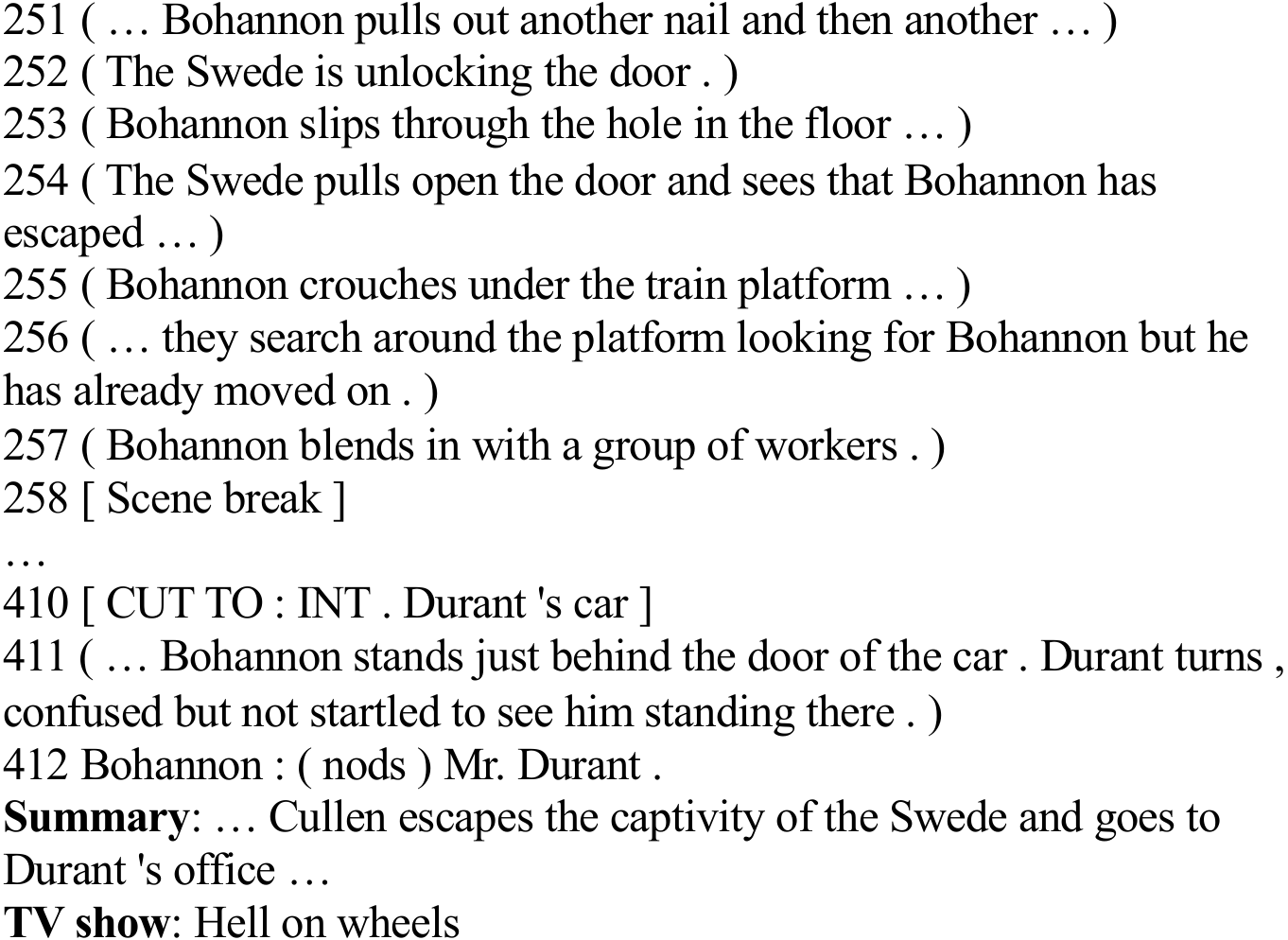}
    \end{subfigure}
    \caption{Two excerpts from \dataset showing that generating summaries from TV show transcripts requires drawing information from a wide range of the input transcripts. We only show lines in the transcripts that are closely related to the shown parts of summaries. The number at the beginning of each line is the line number in the original transcript. For the first instance, we omit a few lines containing clues about the doctor taking pictures of the mansion at different times due to space constraints.}
    \vspace{-0.5em}
    \label{fig:long_range_example}
\end{figure*}

A related challenge is the need to understand the context in which characters' utterances are situated. In the example, the recap describes %
four characters taking sides regarding %
playing a board game. %
The transcript expresses the characters' sentiments through their interactions with one another. The conflict does not occur until Sheldon proposes to ``stay here and play Lord of the Rings Risk'', and it becomes more apparent when Penny mentions she does not want to play the board game. Given the context, Leonard's series of yes and no responses to Penny's questions is largely due to the awkward situation, and it actually shows that he is happy playing the game as he and Sheldon are doing so %
at the beginning of the scene. Similarly, Amy mentions their previous agreement with Sheldon as a way of politely declining Sheldon's plan. The sentiments of characters are not necessarily easily discernible from their utterances but rather must be inferred using context and knowledge about the characters. 

Another challenge in \dataset is the need to draw information from a wide range of the input transcripts, which arises for two primary reasons. 
First, there are many utterances that serve a purpose other than driving the plot forward. 
They may help to develop characters or character relationships, or to add humor or suspense. 
These lines enrich the narrative but their information content is  often omitted from the summaries. 
For example, in the first instance in Figure~\ref{fig:long_range_example}, we show key lines from the transcript that pertain to the excerpt of the summary. There are many other lines between the lines shown, which are conversations between the doctor and other characters. 
This property necessitates the models' ability to correctly attend to major events across the transcript when generating summaries. 
The pattern %
can also be observed in Table~\ref{tab:dataset_stats} through the differences between the number of unique characters in recaps and  transcripts. More than half of the characters in the transcripts are not contained in the recaps. 

The second reason why information needs to be combined across wide ranges of the input relates to scene breaks and multiple plots. As a TV show often narrates a few plots in parallel, scene breaks are used to separate the stories. The discontinuity sometimes requires models to connect subplots hundreds of lines apart. For example, for the second instance in Figure~\ref{fig:long_range_example}, the show uses scene breaks to express what is happening when Cullen Bohannon escapes from the Swede, which is why there are almost two hundred lines between Cullen Bohannon's escape and his appearance at Durant's office.

\section{Approaches}
In this section, we describe modeling approaches that we benchmark on \dataset. We note that since the meaning of sentences in transcripts is highly context-dependent, extractive summarization approaches are not expected to be useful for this dataset. 
We report the results from nearest neighbor-based extractive summarizers mostly for characterizing the dataset.

\subsection{Neural Models}

We use transformer based sequence-to-sequence architectures \cite{attenion_is_all_you_need}. 
Because transcripts are quite long, 
We limit the number of encoder hidden vectors that are used in the decoder's attention mechanism. To do so, when encoding transcripts, we first append a special token ``[EOS]'' to each line of the transcript, and then linearize the transcript.
We then only feed the vectors representing these special tokens to the decoder. We use the Longformer  \cite{beltagy2020longformer} as our encoder architecture, and set the ``[EOS]'' tokens to use global attention.
For our decoders, we use the standard transformer architecture.

\subsection{Nearest Neighbor Models}

We consider two metrics when finding nearest neighbors: BM25 \cite{robertson1995okapi} (a popular metric for information retrieval),
and ROUGE scores \cite{lin-2004-rouge}. We use ROUGE scores as they are used for evaluation, and we use BM25 because it is designed for retrieving long documents whereas ROUGE scores are not. When using ROUGE scores, we use the average of ROUGE-1, ROUGE-2, and ROUGE-L. We consider three types of nearest neighbor search: transcript-to-transcript, recap-to-transcript, and recap-to-recap.

\paragraph{Recap-to-transcript (NNM-r2t).} We use each sentence in the recap as queries and the lines in the corresponding transcript as candidates. The generation is formed by the nearest neighbors for each sentence. We use BM25 or ROUGE scores as the metric. This method can serve as an oracle result for an extractive summarization system, showing roughly how much information can be extracted at the utterance level from the source transcript.

\paragraph{Transcript-to-transcript (NNM-t2t).} We use the transcripts in the test sets as queries, the transcripts in the training sets as candidates, and then find the nearest neighbors using BM25. The generations are the corresponding recaps.
This baseline measures the similarity of instances between training and test splits.

\paragraph{Recap-to-recap (NNM-r2r).} This setting is similar to the ``transcript-to-transcript'' setting, but we use recaps for both queries and candidates, and we use ROUGE and our proposed entity-centric scores (see Sec. \ref{sec:entity_metrics} for more details) as the metric. When using the entity metrics, we use the average of the 4 metric scores. This is an oracle baseline of the ``transcript-to-transcript'' setting and also measures the similarity of the splits.

\subsection{Hybrid Models}

As content selection has been shown to be helpful in prior work \cite{gehrmann-etal-2018-bottom,liu2018generating}, we use the ``recap-to-transcript'' nearest neighbor model and BM25 as the metric to select the most salient content from transcripts, and then apply neural models to the selected content when performing generation. As these methods combine nearest neighbor models and neural models, we refer to them as hybrid models.

In particular, for each sentence in the recap, we find the top three most similar lines in the transcript, include two extra lines that come before or after the selected lines as context, and also include a line that is retrieved by using the whole recap.
As the selected content is significantly shorter than the original transcript, it allows us to use pretrained models.\footnote{After the selection steps, the average number of tokens of the transcripts for \foreverdreamshort and \tvmegasiteshort reduces to 1138.9 and 3252.7 respectively.} Therefore, in this setting, we fine-tune a pretrained BART-large model \cite{lewis-etal-2020-bart}. We note that as the nearest neighbor models rely on the gold standard recaps, this hybrid model demonstrates an approximate upper bound on performance when using powerful content selectors.\footnote{We use the maximum sequence length of 1024 (i.e., we truncate the input sequences if they are longer than 1024) for BART-large due to computational constraints. }

To establish a non-oracle baseline, we train neural models to predict the selected lines, and then fine-tune BART-large models on the predicted lines. Details of the architecture for this component, which we call our ``neural content selector'', are in the appendix.

\section{Experiments}

\begin{table*}[t]
    \centering
    \small\setlength{\tabcolsep}{4pt}
\begin{tabular}{|l|ccccc|ccccc|}
\hline%
& \multicolumn{5}{c|}{ Generic Metrics} & \multicolumn{5}{c|}{ Entity Metrics} \\
& BLEU & R1 & R2 & RL & avg. & BoC-p & BoC-r & BoR-p & BoR-r & avg. \\
\hline
\multicolumn{11}{c}{ \foreverdream } \\\hline
NNM-r2t (oracle, BM25) & 3.4 & 34.3 & 6.6 & 29.6 & 18.5 & 70.5 & 61.9 & 36.4 & 16.1 & 46.2 \\
NNM-r2t (oracle, RG) & 3.9 & 34.8 & 8.5 & 31.5 & 19.7 & \bf 76.7 & 63.3 & \bf 46.5 & 21.3 & 52.0 \\
NNM-r2r (oracle, RG) & \bf 9.9 & \bf 38.8 & \bf 11.5 & \bf 33.9 & \bf 23.5 & 50.6 & 51.4 & 24.6 & 26.8 & 38.4 \\
NNM-r2r (oracle, Entity Metrics) & 5.5 & 31.1 & 6.8 & 27.1 & 17.6 & 58.6 & \bf 79.6 & 26.4 & \bf 43.7 & \bf 52.1 \\
NNM-t2t & 7.9 & 31.3 & 7.8 & 27.4 & 18.6 & 56.5 & 59.2 & 28.2 & 29.4 & 43.3 \\\hline
Neural model & 2.6 & 25.9 & 4.2 & 23.8 & 14.1 & 54.7 & 38.5 & 22.8 & 15.1 & 32.8 \\
Hybrid model & 2.4 & 25.3 & 3.9 & 23.1 & 13.7 & 61.2 & 51.4 & 29.8 & 23.6 & 41.5 \\
Hybrid model (oracle) & 3.0 & 26.4 & 5.0 & 23.3 & 14.4 & 70.0 & 57.8 & 36.9 & 29.1 & 48.5 \\
\hline
\multicolumn{11}{c}{ \tvmegasite } \\\hline
NNM-r2t (oracle, BM25) & 6.7 & 45.0 & 10.2 & 43.0 & 26.2 & 82.5 & \bf 80.4 & 57.7 & 18.1 & 59.7 \\
NNM-r2t (oracle, RG) & \bf 8.5 & 44.1 & 11.7 & 42.4 & 26.7 & 85.2 & 76.8 & \bf 61.2 & 16.9 & 60.0 \\
NNM-r2r (oracle, RG) & 7.9 & \bf 49.0 & 11.6 & \bf 46.9 & \bf 28.9 & 59.2 & 59.0 & 29.5 & 29.9 & 44.4 \\
NNM-r2r (oracle, Entity Metrics) & 4.9 & 42.8 & 8.8 & 40.4 & 24.2 & 60.8 & 81.7 & 26.0 & \bf 37.5 & 51.5 \\
NNM-t2t & 6.2 & 43.2 & 8.6 & 41.4 & 24.9 & 63.2 & 69.3 & 31.8 & 35.3 & 49.9 \\\hline
Neural model & 7.9 & 42.9 & 11.9 & 41.6 & 26.1 & \bf 86.1 & 48.7 & 48.9 & 22.3 & 51.5 \\
Hybrid model & 5.5 & 38.8 & 10.2 & 36.9 & 22.8 & 84.5 & 57.2 & 51.0 & 29.3 & 55.5 \\
Hybrid model (oracle) & 8.9 & 42.1 & \bf 11.9 & 40.9 & 25.9 & 84.0 & 69.5 & 56.4 & 36.8 & \bf 61.7 \\
\hline
\end{tabular}
    \caption{
    Results on the \dataset test sets. 
    BLEU, R1, R2, and RL are BLEU and ROUGE scores between model generated and reference recaps. 
    Bo\{C,R\}-\{p,r\} are precision and recall for bag of characters and bag of character relations, respectively. The highest numbers for each dataset in each column are in bold. 
    }
    \vspace{-0.5em}
    \label{tab:result}
\end{table*}
\subsection{Setup}
\label{sec:setup}

\label{sec:entity_metrics} We report BLEU \cite{papineni-etal-2002-bleu}, ROUGE-1 (R1), ROUGE-2 (R2), and ROUGE-L (RL). We report the average of these four metrics as it generally shows the semantic similarities between generations and references.
We will refer to these metrics as generic metrics as they treat each word equally. 

As characters are fundamental to TV show plots, we believe the quality of plot summaries also depends on including the right characters. %
To take this factor into account, we compute several bag of character (BoC) metrics based on the fraction of the overlapping characters between generated and gold standard recaps. Formally, we define the BoC precision to be
\begin{equation}
    \frac{\vert f(\text{generation})\&f(r)\vert}{\vert f(\text{generation})\vert}
\end{equation}
\noindent where $f$ is a function that extracts the bag of characters from some text, where we perform string matching based on the character names that are automatically extracted during dataset construction (see Sec. \ref{sec:dataset_construct}), $\&$ computes the intersection of two bags, $\vert\cdot\vert$ returns the size of its inputs, %
and $r$ is the gold standard recap. Similarly, we define the BoC recall to be
\begin{equation}
    \frac{\vert f(\text{generation})\&f(r)\vert}{\vert f(r)\vert}
\end{equation}
Since BoC does not consider relations between characters, we also report bag of character relations (BoR) metrics based on the cooccurrence of character pairs. We assume two characters are related when they appear in the same sentence.
After obtaining the character relations from the gold standard recaps and the generations, we compute recall and precision against the recaps following the same approach as BoC. We note that the extracted relations are non-directional, and BoR does not consider frequency of the cooccurrences. We also report the averages of both precisions and recalls from both the BoC and BoR metrics.

More details about hyperparameters are in the appendix.

\subsection{Results}
\label{sec:main_results}

We report test results for \foreverdreamshort and \tvmegasiteshort in Table~\ref{tab:result}.
Our findings for the nearest neighbor models are as follows:

\begin{enumeratesquish}
\item
We find that the nearest neighbor models give strong performance on our dataset. In particular, NNM-r2t shows the best performance in general. 
This demonstrates that there is still room for improving the ability of our neural models to extract the most useful information from transcripts, 
suggesting that improved transcript modeling may be a fruitful research direction for these datasets.
\item
We observe that NNM-r2r exhibits different strengths when based on different metrics, for example, using ROUGE scores will lead to results favorable to generic metrics.
\end{enumeratesquish}

As for the results involving neural models, our findings are as follows:

\begin{enumeratesquish}
\item
The neural model shows strong performance in generic semantic matching but it is relatively weak in entity metrics compared to the non-oracle baselines. 
(see the appendix for more discussion).
\item
The hybrid model is better than the neural model in terms of generating character mentions and relations. With the help of the oracle content selector, the hybrid model improves significantly in both semantic matching and entity-related metrics, showing that future research may find improvement by designing better content selectors.
\end{enumeratesquish}

\section{Analysis}

\begin{table}[t]
    \centering\small
    \begin{tabular}{|l|c|c|}\hline %
        & Generic & Entity \\
        \hline %
        \multicolumn{3}{c}{\foreverdream} \\
        \hline %
        \foreverdreamshort Only & 16.5 & 47.3  \\
        \tvmegasiteshort + \foreverdreamshort & 16.9 & 50.1 \\
        \hline %
        \multicolumn{3}{c}{\tvmegasite} \\\hline %
        \tvmegasiteshort Only & 25.9 & 61.7 \\
        \tvmegasiteshort + \foreverdreamshort & 23.2 & 58.0 \\
        \hline %
    \end{tabular}
    \caption{Results of 
    the 
    oracle hybrid model comparing training on both datasets (\tvmegasiteshort + \foreverdreamshort) to training on the in-domain dataset only. The metrics are averaged scores of the generic and entity metrics. Training on both datasets helps for \foreverdreamshort but hurts for \tvmegasiteshort.} 
    \label{tab:effect_combine}
\end{table}

\begin{table}[t]
    \centering\small
    \begin{tabular}{|l|c|c|}\hline
& Predicates & Character Relation \\\hline
NNM-t2t & 1.6$\pm$0.8 & 2.1$\pm$1.1 \\
Hybrid model & 2.3$\pm$0.9 & 2.0$\pm$1.0 \\
Hybrid model (oracle) & 2.4$\pm$1.0 & 2.4$\pm$1.0 \\\hline
\end{tabular}
    \caption{Human evaluation results. We report the average scores and their corresponding standard deviations for questions on predicate match and character relation similarity. }
    \vspace{-0.7em}
    \label{tab:human_eval}
\end{table}

\begin{table*}[t]
    \centering\small
    \begin{tabular}{|p{0.1\textwidth}|p{0.8\textwidth}|}\hline
Reference & The remains of two witches , one of which is from the Salem witch trials from the 1600s and the other a modern day Wiccan , are discovered in the remains of a burnt out cabin . Booth and Brennan investigate the world of Wicca , including discovering the Wiccan group of which the victim was a part . Hodgins and Angela wind up in jail after some reckless driving and have to work the case from the jail cell . After spending quality time together , they realize they are still in love . Hodgins subsequently proposes to Angela and they are married by the judge who hears their case . \\\hline
Neural model & Booth and Brennan are called to investigate when they are found dead in the death of a young woman who is found in to investigate . Meanwhile , Brennan and Booth are found at the victim 's death of an old friend , but the team must find out to investigate the team up with the case . The team investigates a young man who was killed when they discover that the victim was killed . The victim was not have been retrieve her father , Booth and Angela and Booth 's father 's death . \\\hline
Hybrid model & While the team investigates the death of a 40-year - old woman , who was found buried in a rock quarry . They discover that the woman 's feet were curled after she was buried , and that the bones were de - fleshed prior to her death . Meanwhile , Hodgins and Angela are in jail . Hodgins tells Angela that he 's a witch , but Angela tells Hodgins that she 's not a witch . The team finds out that the victim 's sister , Cheri , was also buried in the quarry . \\\hline
Hybrid model (oracle) & While the team investigates the death of a woman found buried in the woods . They discover that the victim was a Wiccan , and that she may have been a victim of a ritual that was used during the Salem Witch Trials . They also find that the woman was wearing red slippers and that her feet curled up after she was dead . Meanwhile , Hodgins and Angela are in a jail cell , and they are having a hard time adjusting to their new life in the city . The case is complicated by the fact that the body of the woman who was found is a young girl . \\\hline
\end{tabular}
    \caption{Generation samples from \foreverdream. The instance is from the TV show ``Bones''.}
    \vspace{-0.5em}
    \label{tab:gen_example}
\end{table*}

\section{Effect of Combining \foreverdreamshort and \tvmegasiteshort}
\label{sec:combine_datasets}

We study the effect of transfer learning using these two resources. When doing so, we use the training and development sets constructed from both resources, and at test time, we evaluate models on the official test splits. We experiment with the oracle hybrid model and report results in Table \ref{tab:effect_combine}. In general, we find that extra training data helps \foreverdreamshort. We hypothesize that this is due to the relatively small size of \foreverdreamshort. However, for \tvmegasiteshort, training on \foreverdreamshort harms performance, which is likely because of the larger training set size for \tvmegasiteshort and the differences between the two resources.

\subsection{Human Evaluation}

We conduct human evaluations for three models: NNM-t2t, hybrid model, and hybrid model (oracle). To evaluate two key aspects of \dataset, namely events and characters relationships, we ask human annotators two questions.  The first is ``Do the predicates in the generation match the predicates in the reference?''\footnote{By ``predicate'' here we mean the part of a sentence or clause containing a verb and stating something about the subject (e.g., ``went home'' in ``John went home'').} The second is ``When multiple characters are mentioned as being related in some way in the generated recap, are those same characters mentioned as being related in some way in the reference?'' We disregard the subjects in the first question because the second question involves evaluating characters and we want the two questions to focus on different aspects to maximize the efficiency of human annotations. %
Ratings are on a 1-5 scale with 5 indicating a perfect match. 
We randomly picked instances from the \foreverdreamshort test set. We (the authors) annotated 120 instances in total for each question. 

After dropping 2 invalid annotations for the second question (as there may not be multiple characters mentioned), we summarize results in Table \ref{tab:human_eval}. While trends for the model performance on character relations are generally similar to our observations in Table \ref{tab:result}, the results for predicate match are very different for NNM-t2t. This is likely because the first question is about predicates disregarding the correctness of the participants. 
We also want to highlight that compared to the oracle hybrid model, the non-oracle one shows competitive performance on predicate match but is less close in terms of generating correct character relations, showing future opportunities for improving this model.

\subsection{Generation Samples}

In Table \ref{tab:gen_example}, we show generation samples for the following models: the neural model, the hybrid model, and the oracle hybrid model.
The neural model manages to fit most of the character names from the reference into the generation. The generation shares similar topics with the reference, but compared to the hybrid models it lacks specifics. This matches our observations from the automatics metrics where the neural model performs better on the generic metrics but worse on the entity metrics on the non-anonymized datasets. We hypothesize that this is caused by the difficulty of modeling long-form text. 

In the output of the non-oracle hybrid model, many facts that are not mentioned in the reference are actually from the transcript. For example, ``40-year-old woman'' and ``de-fleshed prior to her death'' are in the transcript. Despite containing many specifics, the generation misses a few important details, such as the absence of mentioning main characters involved (i.e., Brennan and Booth). It also has incorrect facts. For example, according to the transcript, there are rocks at the scene, but the model describes the setting as a rock quarry. Compared to the other three models, the generation from the oracle hybrid model is the most faithful,
although there are still incorrect facts (e.g., ``... and they are having a hard time adjusting to their new life in the city.''). The differences between the oracle and non-oracle hybrid model suggest that future research can focus on improving models' capabilities of doing content selection. As both oracle and non-oracle hybrid models suffer from generating incorrect facts, %
faithfulness in generation is also an important future research direction. 

\section{Conclusion}

We construct \dataset, which contains pairs of TV show transcripts and recaps. We qualitatively analyze the challenging aspects of our dataset. We propose two entity-centric metrics to evaluate generated summaries with one focusing on character overlap and the other focusing on overlap of character pairs. Empirically, 
we benchmark several neural models and nearest neighbor models for characterizing our datasets, finding that an oracle extractive summarizer gives the strongest performance according to the automatic metrics. Human evaluations show that the non-oracle hybrid model is competitive at generating faithful topics. Qualitative analysis shows that the hybrid model can benefit from better content selectors and both oracle and non-oracle hybrid models suffer from generating inaccurate details, highlighting several directions for future research. 

\section*{Acknowledgments}
We wish to thank The TV MegaSite, Inc.~and Forever Dreaming for allowing us to use and redistribute their data for research purposes. This work was supported in part by a Google Fellowship to M. Chen.

\bibliography{anthology,custom}
\bibliographystyle{acl_natbib}

\clearpage

\appendix

\section{Hyperparameters}

We set the maximum sequence length to be 14336 for the encoder and 1024 for the decoder. We use byte-pair encoding \cite{sennrich-etal-2016-neural} with approximately 10k vocabulary size. 
We use a 1-layer encoder and a 12-layer decoder with 1024 hidden units unless otherwise specified. We use an effective batch size of 200, and train the models for 50 epochs. During training, we perform early stopping on the development sets based on perplexities. During testing, we use beam search with trigram blocking \cite{paulus2018a} and a beam size of 5.

For the neural content selector, we use a 3-layer longformer encoder followed by a 2-layer feedforward network with GELU activations \cite{hendrycks2016gelus}. We perform early stopping based on F1 scores on the development sets, where the threshold is chosen by averaging over the oracle thresholds for each instance. When selecting content, we use the threshold chosen based on the development set and ensure that no less than 10\% of lines for each transcript are selected. The model achieves test performance (F1 scores) of 19.0 on \foreverdreamshort, 19.2 on anonymized \foreverdreamshort, 41.5 on \tvmegasiteshort, and 40.1 on anonymized \tvmegasiteshort.

\section{Anonymized \dataset}

As plots for TV shows are typically about a limited number of characters, models trained on \dataset may focus on those characters and their typical behaviors rather than the actual 
actions taking place in the input transcripts. To eliminate this effect, we create an anonymized version of \dataset by replacing character names with random character IDs. We ensure that the IDs of particular characters in different episodes are randomly assigned (i.e., IDs are not consistent across episodes). 

\begin{figure}[t]
    \centering
    \includegraphics[scale=0.5]{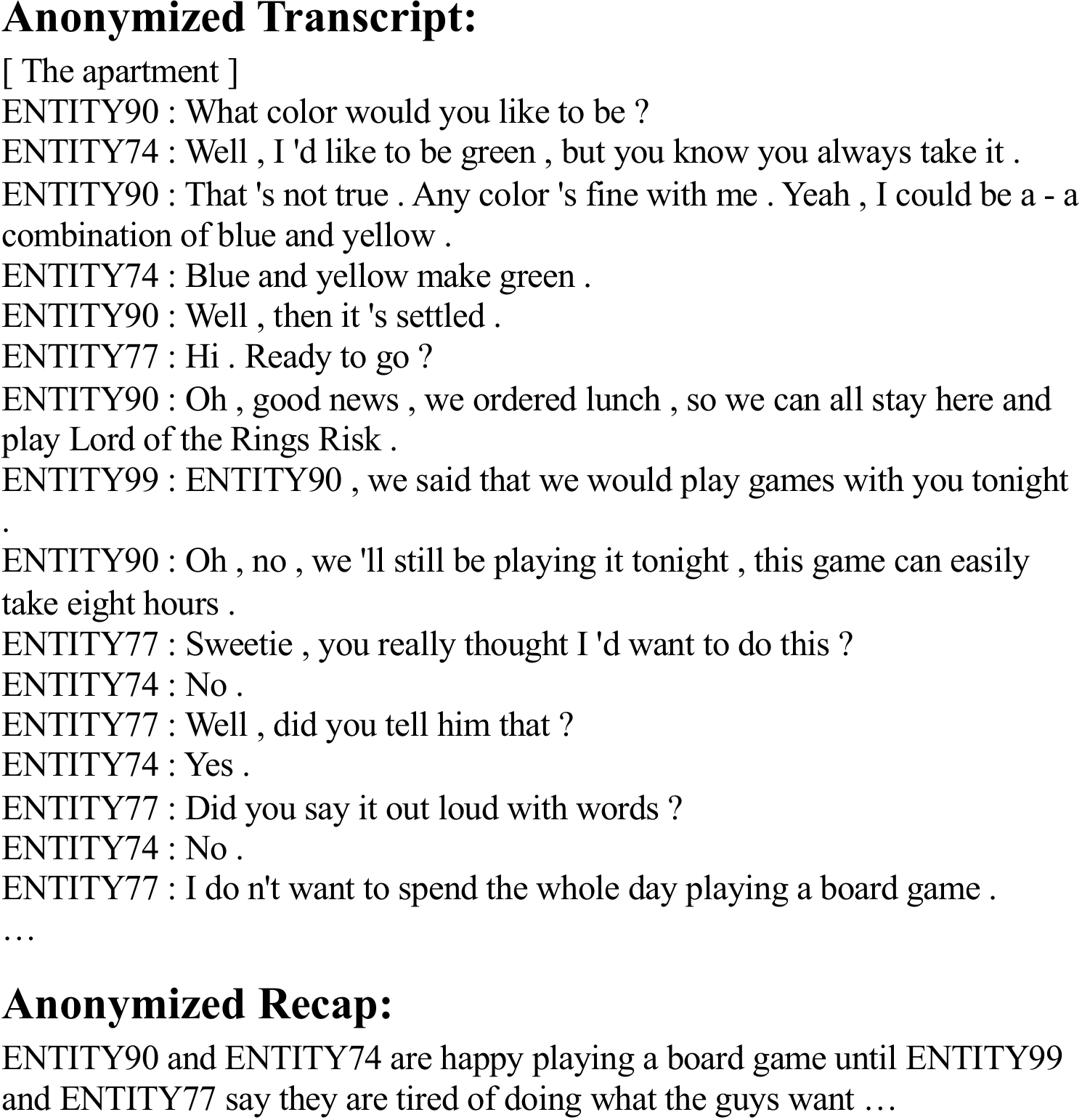}
    \caption{An excerpt from anonymized \dataset that corresponds to the instance in the Figure 1 in the main text. Character names are replaced with IDs that are permuted across episodes.} %
    \label{fig:example}
\end{figure}

Figure~\ref{fig:example} shows an example from anonymized \dataset. Anonymized question answering datasets have also been created out of  similar concerns to those just described \cite{teaching2015hermann}.

\section{Results for the Anonymized Datasets}

\begin{table*}[t]
    \centering
    \small\setlength{\tabcolsep}{4pt}
\begin{tabular}{|l|ccccc|ccccc|}
\hline%
& \multicolumn{5}{c|}{ Generic Metrics} & \multicolumn{5}{c|}{ Entity Metrics} \\
& BLEU & R1 & R2 & RL & avg. & BoC-p & BoC-r & BoR-p & BoR-r & avg. \\
\hline
\multicolumn{11}{c}{ Anonymized \foreverdream } \\\hline
NNM-r2t (oracle, BM25) & 3.5 & 34.5 & 6.8 & 30.0 & 18.7 & 70.4 & 60.4 & 37.5 & 16.7 & 46.2 \\
NNM-r2t (oracle, RG) & 4.0 & \bf 34.7 & 8.5 & \bf 31.4 & 19.7 & \bf 76.8 & \bf 63.4 & \bf 49.1 & 22.6 & \bf 53.0 \\
NNM-r2r (oracle, RG) & \bf 7.9 & 34.3 & \bf 9.1 & 30.1 & \bf 20.4 & 5.4 & 6.3 & 0.2 & 0.1 & 3.0 \\
NNM-t2t & 6.0 & 26.2 & 6.0 & 23.0 & 15.3 & 21.5 & 6.6 & 5.0 & 0.2 & 8.3 \\\hline
Neural model & 2.6 & 28.6 & 4.6 & 25.1 & 15.2 & 65.0 & 57.7 & 27.9 & \bf 30.6 & 45.3 \\
Hybrid model & 2.3 & 23.1 & 3.9 & 20.6 & 12.5 & 12.2 & 2.3 & 0.3 & 0.0 & 3.7 \\
Hybrid model (oracle) & 2.9 & 26.0 & 5.0 & 22.2 & 14.0 & 33.9 & 8.8 & 3.6 & 0.6 & 11.7 \\
\hline
\multicolumn{11}{c}{ Anonymized \tvmegasite } \\\hline
NNM-r2t (oracle, BM25) & 6.9 & \bf 45.0 & 10.2 & \bf 42.9 & 26.2 & 82.6 &\bf 80.5 & 58.9 & 20.7 & 60.7 \\
NNM-r2t (oracle, RG) & \bf 8.7 & 44.1 &  \bf 11.7 & 42.3 & \bf 26.7 & 85.3 & 76.7 & \bf 61.8 & 19.3 & 60.8 \\
NNM-r2r (oracle, RG) & 6.0 & 42.8 & 9.3 & 41.1 & 24.8 & 46.3 & 14.7 & 3.8 & 0.6 & 16.3 \\
NNM-t2t & 4.4 & 26.2 & 6.0 & 23.0 & 14.9 & 47.7 & 15.2 & 3.8 & 0.5 & 16.8 \\\hline
Neural model & 7.1 & 41.6 & 11.6 & 40.4 & 25.2 & \bf 86.8 & 53.6 & 32.0 & 15.2 & 46.9 \\
Hybrid model & 6.2 & 37.7 & 9.3 & 36.4 & 22.4 & 82.5 & 62.3 & 47.4 & 30.2 & 55.6 \\
Hybrid model (oracle) & 6.1 & 38.9 & 10.1 & 37.6 & 23.2 & 84.3 & 68.1 & 55.6 & \bf 38.8 & \bf 61.7 \\
\hline
\end{tabular}
    \caption{
    Results on the anonymized \dataset test sets.
    BLEU, R1, R2, and RL are BLEU and ROUGE scores between model generated and reference recaps. 
    Bo\{C,R\}-\{p,r\} are precision and recall for \emph{Bag of Characters} and \emph{Bag of Character Relations}, respectively. The highest numbers for each dataset in each column are in bold. 
    }
    \label{tab:anon_result}
\end{table*}

In Table~\ref{tab:anon_result}, it is interesting to observe the performance differences of the nearest neighbor models between the anonymized and non-anonymized datasets. The gaps show that the anonymization does not lead to much difference regarding the similarities between recaps and transcripts, but it makes correlations among recaps and transcripts much weaker especially for those entities.

\section{Effect of Anonymization}
\label{sec:effect_anonym}
\begin{table}[t]
    \centering\small
    \begin{tabular}{|l|c|c|}\hline
        Fraction & \tvmegasiteshort & Anonymized \tvmegasiteshort \\\hline
        All &  61.7 & 61.7 \\
        80\% & 19.1 & 25.5 \\
        60\% & 11.0 & 17.0 \\
        \hline
    \end{tabular}
    \caption{Average scores of entity metrics computed on various subsets of entities, dropping the most common entities when forming subsets. For example, the ``80\%'' row corresponds to omitting the most frequent 20\% of entities in each TV show. Results are based on the oracle hybrid model.}
    \label{tab:effect_anonym}
\end{table}

We study the effect of anonymization by investigating performance on rare entities. To do so, we first compute entity frequencies for each TV show 
from the training set, rank the entities by their frequencies, pick the rare entities according to the rank, and evaluate performance for the selected entities. We summarize the results in Table \ref{tab:effect_anonym}. We find that models trained on the anonymized \tvmegasiteshort dataset give better performance on rare entities, suggesting that anonymization helps in modeling rare entities. The fact that the two models have the same performance in the ``all'' setting shows that anonymization also makes the learning of common entities harder, matching our expectations.

\section{Effect of Copy Mechanism}
\label{sec:effect_copy}

\begin{table}[t]
    \centering\small
    \begin{tabular}{|l|c|c|}\hline
         & Generic & Entity \\\hline
        \multicolumn{3}{c}{\foreverdream} \\\hline
        w/o copy mechanism & 12.4 & 29.3 \\
        w/ copy mechanism & 12.6 & 27.1 \\\hline
    \end{tabular}
    \caption{Comparing models with and without the copy mechanism on \foreverdream.}
    \label{tab:copy_mechanism}
\end{table}

We report results on \foreverdream in Table~\ref{tab:copy_mechanism} comparing models with and without the copy mechanism. We note that models used in this table use 6-layer decoders with 512 hidden units, so the results are not directly comparable to other results. From the results in Table~\ref{tab:copy_mechanism}, we find that the copy mechanism helps tremendously on the anonymized dataset, but gives mixed results on the non-anonymized dataset. This is likely due to the fact that for the anonymized dataset, there is not enough training data for the character ID embeddings, and the copy mechanism helps to reduce the required supervision. While there may be better ways of handling the character IDs that may avoid this issue (e.g., sampling IDs from exponential-like distributions rather than uniform distribution), we leave this for future research.  
However, this benefit does not hold for the non-anonymized dataset as the models are able to exploit more information when learning character name embeddings by having access to the character names. 

\section{Effect of Combining \foreverdreamshort and \tvmegasiteshort}

\begin{table}[t]
    \centering\small
    \begin{tabular}{|l|c|c|}\hline %
        & Generic & Entity \\
        \hline %
        \multicolumn{3}{c}{Anonymized \foreverdream} \\
        \hline %
        Anonymized \foreverdreamshort Only & 13.7 & 11.3 \\
        Anonymized (\tvmegasiteshort + \foreverdreamshort) & 17.1 & 52.9 \\
        \hline %
        \multicolumn{3}{c}{Anonymized \tvmegasite} \\\hline %
        Anonymized \tvmegasiteshort Only & 23.2 & 61.7 \\
        Anonymized (\tvmegasiteshort + \foreverdreamshort) & 22.7 & 59.8 \\
        \hline %
    \end{tabular}
    \caption{Results of 
    the 
    oracle hybrid model comparing training on both datasets (\tvmegasiteshort + \foreverdreamshort) to training on the in-domain dataset only. The metrics are averaged scores of the generic and entity metrics. Training on both datasets helps for \foreverdreamshort but hurts for \tvmegasiteshort.} 
    \label{app_tab:effect_combine}
\end{table}

In Table~\ref{app_tab:effect_combine}, it is interesting to see that the anonymized \foreverdream benefits greatly from additional training data, supporting our previous hypothesis that the copy mechanism helps to reduce the amount of required supervision.

\end{document}